\documentclass{article}
\usepackage{ijcai23}

\usepackage{times}
\usepackage{soul}
\usepackage{url}
\usepackage[hidelinks]{hyperref}
\usepackage[utf8]{inputenc}
\usepackage[small]{caption}
\usepackage{graphicx}
\usepackage{amsmath}
\usepackage{amsthm}
\usepackage{booktabs}
\usepackage[switch]{lineno}

\usepackage{color}
\usepackage{subfigure}
\usepackage{multirow}
\usepackage{amssymb}
\usepackage{colortbl}
\usepackage{arydshln,xcolor,array}
\usepackage{comment}
\usepackage{enumerate}
\usepackage{float}
\usepackage[ruled,linesnumbered]{algorithm2e}
\definecolor{edit1}{rgb}{0.3,0.3,0.8}
\title{BPNet: Bézier Primitive Segmentation on 3D Point Clouds}

\author{
Rao Fu$^{1,2}$\and
Cheng Wen$^3$\and
Qian Li$^{1,}$\thanks{Corresponding author}\and
Xiao Xiao$^4$\And
Pierre Alliez$^1$
\affiliations
$^1$Inria, France\\
$^2$Geometry Factory, France\\
$^3$The University of Sydney, Australia\\
$^4$Shanghai Jiao Tong University, P. R. China\\
\emails
\{rao.fu, qian.li, pierre.alliez\}@inria.fr,
cwen6671@uni.sydney.edu.au,
xiao.xiao@sjtu.edu.cn
}
\begin{document}

\maketitle

\begin{abstract}

This paper proposes BPNet, a novel end-to-end deep learning framework to learn Bézier primitive segmentation on 3D point clouds. The existing works treat different primitive types separately, thus limiting them to finite shape categories. To address this issue, we seek a generalized primitive segmentation on point clouds. Taking inspiration from Bézier decomposition on NURBS models, we transfer it to guide point cloud segmentation casting off primitive types. A joint optimization framework is proposed to learn Bézier primitive segmentation and geometric fitting simultaneously on a cascaded architecture. Specifically, we introduce a soft voting regularizer to improve primitive segmentation and propose an auto-weight embedding module to cluster point features, making the network more robust and generic. We also introduce a reconstruction module where we successfully process multiple CAD models with different primitives simultaneously. We conducted extensive experiments on the synthetic ABC dataset and real-scan datasets to validate and compare our approach with different baseline methods. Experiments show superior performance over previous work in terms of segmentation, with a substantially faster inference speed.

\end{abstract}

\section{Introduction}
Structuring and abstracting 3D point clouds via segmentation is a prerequisite for various computer vision and 3D modeling applications. Many approaches have been proposed for semantic segmentation, but the finite set of semantic classes limits their applicability. 3D instance-level segmentation and shape detection are much more demanding, while this literature lags far behind its semantic segmentation counterpart. Finding a generalized way to decompose point clouds is essential. For example, man-made objects can be decomposed into canonical primitives such as planes, spheres, and cylinders, which are helpful for visualization and editing. However, the limited types of canonical primitives are insufficient to describe objects' geometry in real-world tasks. We are looking for a generalized way of decomposing point clouds. The task of decomposing point clouds into different geometric primitives with corresponding parameters is referred to as \emph{parametric primitive segmentation}. Parametric primitive segmentation is more reasonable than semantic instance segmentation for individual 3D objects, which unifies the 3D objects in the parametric space instead of forming artificially defined parts. However, the task is quite challenging as 1) there is no exhaustive repertoire of canonical geometric primitives, 2) the number of primitives and points belonging to that primitive may significantly vary, and 3) points assigned to the same primitive should belong to the same type of primitive.

Inspired by the fact that B\'ezier decomposition, where NURBS models can be divided into canonical geometric primitives (plane, sphere, cone, cylinder, etc.) and parametric surfaces into rational B\'ezier patches, we propose to learn B\'ezier decomposition on 3D point clouds. We focus on segmenting point clouds sampled from individual objects, such as CAD models. Departing from previous primitive segmentation, we generalize different primitive types to B\'ezier primitives, making them suitable for end-to-end and batch training. To the best of our knowledge, our method is the only work to learn B\'ezier decomposition on point clouds. To summarize our contributions:
\begin{enumerate}
\item We introduce a novel soft voting regularizer for the relaxed intersection over union (IOU) loss, improving our primitive segmentation results.
\item We design a new auto-weight embedding module to cluster point features which is free of iterations, making the network robust to real-scan data and work for axis-symmetric free-form point clouds.
\item We propose an innovative reconstruction module where we succeed in using a generalized formula to evaluate points on different primitive types, enabling our training process to be fully differential and compatible with batch operations.
\item Experiments demonstrate that our method works on the free-form point clouds and real-scan data even if we only train our model on the ABC dataset. Furthermore, we present one application of B\'ezier primitive segmentation to reconstruct the full B\'ezier model while preserving the sharp features. The code is available at: \url{https://github.com/bizerfr/BPNet}. 
\end{enumerate}

\section{Related Work}

Bézier primitive segmentation involves parametric fitting, instance segmentation, and multi-task learning. We now provide a brief review of these related research areas.

\paragraph{Primitive segmentation.} Primitive segmentation refers to the search and approximation of geometric primitives from point clouds. Primitives can be canonical geometric primitives, such as planes or spheres, or parametric surface patches, such as B\'ezier, BSpline, or NURBS. We can classify primitive segmentation methods into two lines of approaches: geometric optimization and machine learning. Popular geometric optimization-based methods include RANSAC \cite{fischler1981random,schnabel2007efficient}, region growing  \cite{marshall2001robust} and Hough transforms \cite{rabbani2007integrated}. We refer to \cite{kaiser2019survey} for a comprehensive survey. One limitation of geometric optimization-based methods is that they require strong prior knowledge and are hence sensitive to parameters. In order to alleviate this problem, recent approaches utilize neural networks for learning specific classes of primitives such as cuboids \cite{zou20173d,tulsiani2017learning}. The SPFN supervised learning approach \cite{li2019supervised} detects a wider repertoire of primitives such as planes, spheres, cylinders, and cones. Apart from the canonical primitives handled by SPFN, ParSeNet \cite{sharma2020parsenet} and HPNet \cite{yan2021hpnet} also detect open or closed BSpline surface patches. Nevertheless, different types of primitives are treated separately with insufficient genericity. This makes them unsuitable for batch operations, thus suffering long inference times. Deep learning-based methods are less sensitive to parameters but often support a limited repertoire of primitives. Our work extends SPFN, ParSeNet, and HPNet with more general Bézier patches.

\paragraph{Instance segmentation.} Instance segmentation is more challenging than semantic segmentation as the number of instances is not known a priori. Points assigned to the same instance should fall into the same semantic class. We distinguish between two types of methods: proposal-based \cite{yi2019gspn,yang2019learning,engelmann20203d} and proposal-free methods \cite{wang2018sgpn,jiang2020pointgroup,huang2021primitivenet}. On the one hand, proposal-based methods utilize an object-detection module and usually learn an instance mask for prediction. On the other hand, proposal-free methods tackle the problem as a clustering step after semantic segmentation. We refer to a recent comprehensive survey \cite{guo2020deep}. The significant difference between instance segmentation and primitive segmentation is that instance segmentation only focuses on partitioning individual objects where primitive fitting is absent.

\paragraph{Patch-based representations.} Patch-based representations refer to finding a mapping from a 2D patch to a 3D surface. Previous works including \cite{groueix2018papier,yang2018foldingnet,deng2020better,bednarik2020shape} learn a parametric 2D mapping by minimizing the Chamfer distance \cite{fan2017point}. One issue with Chamfer distance is that it is not differentiable when using the nearest neighbor to find matched pairs. We learn the $uv$ mapping instead. Learning $uv$ parameters enables us to re-evaluate points from our proposed generalized B\'ezier primitives, making our training process differentiable and supporting batch operations.  
 
\paragraph{Multi-task learning.} Multi-task learning aims to leverage relevant information contained in multiple related tasks to help improve the generalization performance of all the tasks \cite{zhang2021survey}. Compared to single-task learning, the architectures used for multi-task learning---see, e.g., \cite{zhang2014facial,dai2016instance}---share a backbone to extract global features, followed by branches that transform the features and utilize them for specific tasks. Inspired by \cite{dai2016instance}, we use a cascaded architecture for our joint optimization tasks.

\section{Method}

\begin{figure*}
\centering
\includegraphics[width=17.5cm]{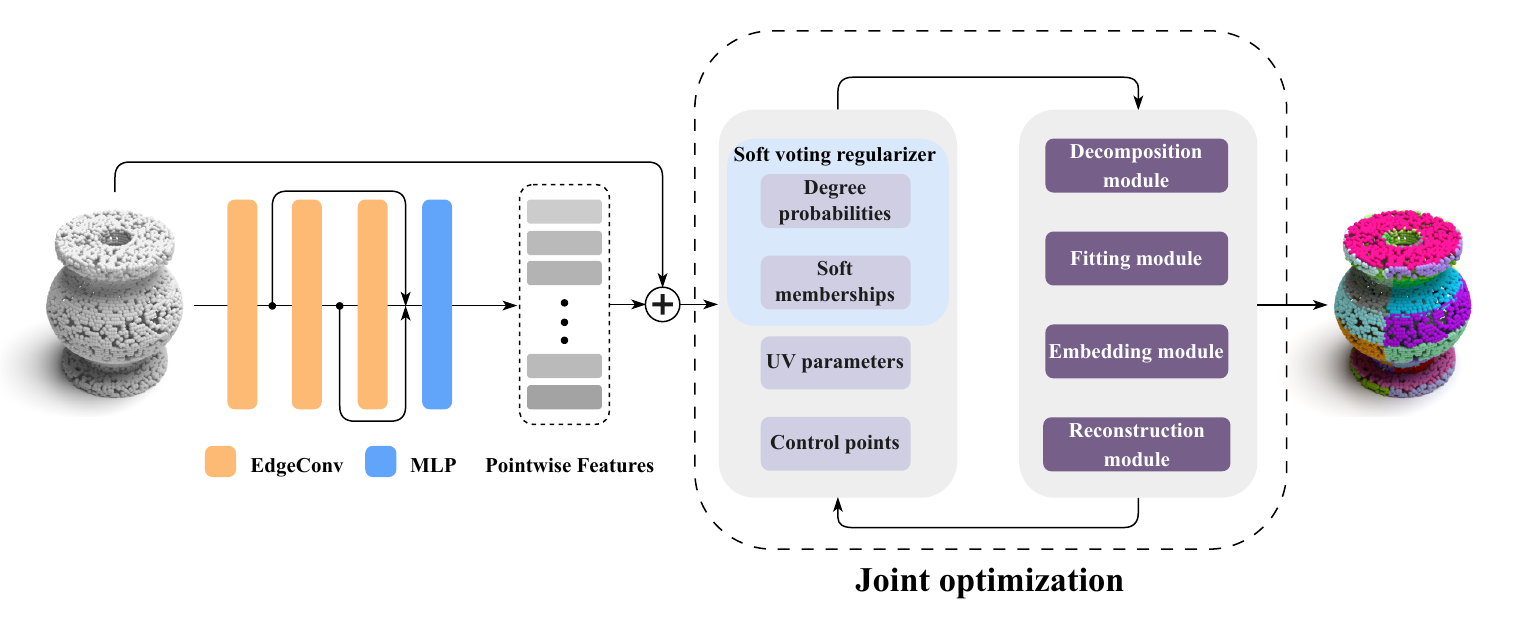}
\caption{Overview of the proposed pipeline. The network takes a point cloud as input. It outputs pointwise features followed by four modules to predict Bézier geometry and topology: (1) The decomposition module decomposes point clouds into multiple patches. (2) The fitting module regresses $uv$ parameters for each point and control points for each Bézier patch. (3) The embedding module clusters pointwise features assigned to the same patch. (4) The reconstruction module re-evaluates the input point clouds from the above predictions.}
\label{fig:pipeline}
\end{figure*}

Figure \ref{fig:pipeline} shows an overview of the proposed neural network. The input to our method is a 3D point cloud $P=\{p_i | 0\leq i \leq N-1\}$, where $p_i$ denotes the point coordinates (with or without normals). The output is the per-point patch labels $\{ P_k | \cup_{k=0} P_k = P\}$, where each patch corresponds to a B\'ezier primitive. The network will also output patch degree ($d_u$-by-$d_v$) and weighted control points $C=\{\mathbf{c}_{kmn} = (x,y,z,w)|0\leq m \leq d_u, 0\leq n \leq d_v, 0 \leq k \leq K-1\} $, where $K$ denotes the number of patches. We constrain the maximum degree to be $M_d*N_d$. We let our network output a maximum number of $K$ Bézier patches for all CAD models, and we use $\hat{K}$ to denote the ground-truth number of patches which is smaller than $K$ and varies for each CAD model. 

\subsection{Architecture}

Our architecture consists of two components: a backbone for extracting features and a cascaded structure for joint optimization. The backbone is based on three stacked \emph{EdgeConv} \cite{wang2019dynamic} layers and extracts a 256D pointwise feature for each input point. Let $\mathbf{P} \in \mathbb{R}^{N \times D_{in}}$ denote the input matrix, where each row is the point coordinates ($D_{in}$ is three) with optional normals ($D_{in}$ is six). Let $\mathbf{X} \in \mathbb{R}^{N \times 256}$ denote the 256D pointwise feature matrix extracted from the backbone. We use a cascaded structure to optimize the per-point degree probability matrix $\mathbf{D} \in \mathbb{R}^{N \times (M_d*N_d) }$, the soft membership matrix $\mathbf{W} \in \mathbb{R}^{N \times K}$, the $UV$ parameter matrix $\mathbf{T} \in \mathbb{R}^{N \times 2}$, and the weighted control points tensor $\mathbf{C} \in \mathbb{R}^{K \times (M_d+1) \times (N_d+1) \times 4}$ jointly. Because $\mathbf{D}$, $\mathbf{W}$, $\mathbf{T}$, and $\mathbf{C}$ are coupled, it is natural to use a cascaded structure to jointly optimize them. Here, the cascaded structure is similar to \cite{dai2016instance}, where the features are concatenated and transformed for different MLP branches.

\subsection{Joint Optimization}

We have four modules: decomposition, fitting, embedding, and reconstruction. They are coupled to optimize $\mathbf{D}$, $\mathbf{W}$, $\mathbf{T}$ and $\mathbf{C}$ jointly by using our proposed four modules.

\subsubsection{Decomposition Module}

\paragraph{Degree classification.}
We use B\'ezier primitive with different degrees to replace classical primitives, including plane, sphere, plane, BSpline, etc. For the sake of the classification of degrees, the straightforward idea would be to use a cross-entropy loss: ${\mathrm{CE}} = -\mathrm{log}(p_t)$, where $p_t$ denotes the possibility of the true degree labels. However, the degree type is highly imbalanced. For example, surfaces of degree type 1-by-1 represent more than $50\%$, while 3-by-2 surfaces are rare. To deal with the imbalance, we utilize the multi-class focal-loss \cite{lin2017focal}: ${\mathrm{FL}} = -(1-p_t)^{\gamma}\mathrm{log}(p_t)$, where $\gamma$ denotes the focusing parameter. Then the degree type classification loss is defined as:
\begin{align}
L_{\mathrm{deg}} = \frac{1}{N}\sum_{i=0}^{N-1} {\mathrm{FL}}(\mathbf{D}_{i,:})
\end{align}

\paragraph{Primitive segmentation.} The output of primitive segmentation is a soft membership indicating per-point primitive instance probabilities. Each element $w_{ik}$ is the probability for a point $p_i$ to be a member of primitive $k$. Since we can acquire pointwise patch labels from our data pre-processing, we use a relaxed IOU loss \cite{krahenbuhl2013parameter,yi2018deep,li2019supervised} to regress the $\mathbf{W}$: 
\begin{small}
\begin{align}
  L_{\text{seg}} = \frac{1}{\hat{K}}\sum_{k=0}^{\hat{K}-1} \left[1 - \frac{ \mathbf{W}_{:,k}^T \mathbf{\hat{W}}_{:,\hat{k}} } {\|\mathbf{W}_{:,k}\|_1 + \|\mathbf{\hat{W}}_{:,\hat{k}}\|_1 - \mathbf{W}_{:,k}^T \mathbf{\hat{W}}_{:,\hat{k}} }\right],
\end{align}
\end{small}
where $\mathbf{W}$ denotes the output of the neural network and $\hat{\mathbf{W}}$ is the one-hot encoding of the ground truth primitive instance labels. The best matching pairs $(k, \hat{k})$ between prediction and ground truth are found via the Hungarian matching \cite{kuhn1955hungarian}. Please refer to \cite{li2019supervised} for more details.

\paragraph{Soft voting regularizer.} Since we learn $\mathbf{D}$ and $\mathbf{W}$ separately, points belonging to the same primitive instance may have different degrees, which is undesirable. To favor degree consistency between points assigned to the same primitive, we propose a soft voting regularizer that penalizes pointwise degree possibilities. We first compute a score for each degree case for all primitive instances by $\mathbf{S} = \mathbf{W}^T\mathbf{D}$, where each element $s_{kd}$ denotes the soft number of points for degree $d$ in primitive instance $k$. We then perform $L_1$-normalization to convert $\mathbf{S}$ into primitive degree distributions $\mathbf{\hat{S}}$:
 \begin{align}
  \mathbf{\hat{S}} = \left[\frac{1}{\sum_{d=0}S_{kd}}\right] \odot \mathbf{S},
\end{align}
where the first term denotes the sum of each column and $\odot$ denotes the element-wise product. Finally, we utilize a focal loss to compute the primitive degree voting loss:
 \begin{align}
  L_{\text{voting}} = \frac{1}{\hat{K}}\sum_{k=0}^{\hat{K}-1} {\mathrm{FL}}({\mathbf{\hat{S}}}_{k,:}), 
\end{align}
where $\mathrm{FL}$ denotes the focal loss. The global loss for the decomposition module is defined as: $  L_{\text{dec}}= L_{\text{deg}} + L_{\text{seg}} + L_{\text{voting}}.$

\subsubsection{Fitting Module}

\paragraph{Parameter regression.} Through B\'ezier decomposition we obtain the ground truth labels for the $(u, v)$ parameters and record all parameters into matrix $\hat{\mathbf{T}}$. We regress the $uv$ parameters using a mean squared error (MSE) loss:
 \begin{align}
  L_{\text{para}}= \frac{1}{N}\sum_{i=0}^{N-1} \|\mathbf{T}_{i,:} - \hat{\mathbf{T}}_{i,:}\|_2^2
\end{align}

\paragraph{Control point regression.} We select a maximum number of primitive instances $K$ for all models. As the ground truth primitive instance $\hat{K}$ varies for each model, we reuse the matching pairs directly from the Hungarian matching already computed in the primitive segmentation step. Note that as the predicted degree $(d_u, d_v)$ may differ from the ground truth $(\hat{d_u}, \hat{d_v})$, we align the degree to compute the loss via a maximum operation as $(max(d_u, \hat{d_u}), max(d_v, \hat{d_v}))$. The network always outputs $(M_d+1) \times (N_d+1)$ control points for each primitive corresponding to the predefined maximum degree in $U$ and $V$ direction, and these control points will be truncated by the aligned degree. Furthermore, if the ground-truth degree is smaller than the prediction, we can pad ``fake'' control points that are zero for the ground-truth patch; otherwise, we just use the aligned degree, which is the maximum of the predicted and the ground truth. Finally, the control point loss is defined as: 
 \begin{align}
  L_{\text{ctrl}}=  \frac{1}{N_{\mathbf{c}}}\sum_{t=0}^{N_{\mathbf{c}}-1} \|\mathbf{c}_t - \hat{\mathbf{c}}_t\|_2^2,
\end{align}
where $\mathbf{c}_t$ and $\hat{\mathbf{c}}_t$ denote the matched control points, and $N_\mathbf{c}$ is the number of matched control point pairs. Finally, we define the $L_{\text{fit}}$ loss as: $L_{\text{fit}} = L_{\text{para}} + L_{\text{ctrl}}.$

\subsubsection{Embedding Module}
We use the embedding module to eliminate over-segmentation by pulling point-wise features toward their center and pushing apart different centers. Unlike ParSeNet and HPNet, 1) we do not need a mean-shift clustering step which is time-consuming; 2) we calculate the feature center in a weighted manner rather than simply averaging. The weights are chosen as $\mathbf{W}$ and will be automatically updated in the decomposition module; 3) $\mathbf{W}$ will be further optimized to improve the segmentation. Moreover, our embedding module is suitable for batch operations even though the number of primitive instances for each CAD model and the number of points for each primitive varies. Otherwise, one has to apply mean-shift for each primitive, which deteriorates timing further.

To be specific, we use $\mathbf{W}$ to weight $\mathbf{X}$ to obtain primitive features for all candidate primitive instances. Then, we reuse $\mathbf{W}$ to weigh all the primitive instance features to calculate a ``soft'' center feature for each point. We favor that each point feature embedding should be close to its ``soft'' center feature, and each primitive instance feature embedding should be far from each other. The primitive instance-wise feature matrix $\mathbf{X}_{\text{ins}}$ is defined as:
 \begin{align}
  \mathbf{X}_{\text{ins}} = [\frac{1}{ \sum_{i=0}^{N-1}w_{ik} }] \odot (\mathbf{W}^T\mathbf{X}),
\end{align}
where each row of $\mathbf{X}_{\text{ins}}$ denotes the instance-wise features for each patch. We then compute the ``soft'' center feature matrix $\mathbf{X}_{\text{center}}$ as: $\mathbf{X}_{\text{center}} = \mathbf{W}\mathbf{X}_{\mathrm{ins}}$, where each row denotes the ``soft'' center for each point. 

Then we define $L_{\text{pull}}$ as:
 \begin{align}
  L_{\text{pull}} = \frac{1}{N}\sum_{i=0}^{N-1} {\mathrm{Relu}}(\|\mathbf{X}_{i,:} - (\mathbf{X}_{\mathrm{center}})_{i,:}\|_2^2 - \delta_{\text{pull}}),
\end{align}
and we define $L_{\text{push}}$ as:
 \begin{align}
 \begin{split}
  L_{\text{push}} = 
  \frac{1}{2K(K-1)}\sum_{k_1<k_2} {\mathrm{Relu}}( \delta_{\text{push}} - \\ \|(\mathbf{X}_{\mathrm{ins}})_{k_1,:} - (\mathbf{X}_{\mathrm{ins}})_{k_2,:}\|_2^2 ).
\end{split}
\end{align}

Finally, the total embedding loss $L_{\text{emb}}$ is defined as: $L_{\text{emb}} = L_{\text{pull}} + L_{\text{push}}$. 

\subsubsection{Reconstruction Module} 
The reconstruction module is designed to reconstruct points from the predicted multiple B\'ezier primitives, i.e., rational B\'ezier patches, and further jointly optimize $\mathbf{W}$. One difficulty is that each CAD model has various numbers of primitives, and the degree of each primitive is also different. Therefore, we seek a generalized formula to support tensor operations on re-evaluating points for a batch of CAD models. 
The straightforward approach would be to compute a synthesizing score for all degree types. Assume the maximum number of primitive instances is $K$, and we have $M_d * N_d$ types of different degrees. The total number of combinations is $K * M_d * N_d$. We define a synthesizing score for each case in Einstein summation form: $(s_w)_{kci} = w_{ik} * s_{kc}$, where $w_{ik}$ denotes the probability of point $p_i$ to belong to primitive instance $k$ and $s_{kc}$ denotes the degree score for degree type $m$-by-$n$ indexed with $c = M * (m - 1) + (n - 1)$ for primitive instance $k$ coming from $\mathbf{S}$. Then, we need to normalize $(s_w)_{kdi}$ such that $\sum_{k, d, i} (s_w)_{kdi} = 1$. Finally, the reconstructed point coordinates $p_i$ are defined as:
 \begin{align}
  \begin{pmatrix}
    x_i'\\
    y_i'\\
    z_i'\\
\end{pmatrix} =  \sum_{k,m,n}(s_w)_{kci}\mathrm{\mathbf{R}}_{kmn}(u_i,v_i),
\end{align}
where parameter $(u_i,v_i)$ for point $p_i$ is shared for all combinations. Such a formulation makes extending the formula in matrix form easy and avoids resorting to loop operations.
However, such an approach is too memory-intensive. We thus truncate the degree from the degree probability matrix by re-defining the Bernstein basis function for degree $d$ as:
 \begin{align}
(B_M)_d^l(t)=
\begin{cases}
\binom{d}{l}t^l(1-t)^{d-l}, & l \le d \\
0, & l > d
\end{cases},
\end{align}
where $0 \le l \le M$, and $M$ is the maximum degree. Then, the reconstructed point coordinates for $p_i$ for a degree $m$-by-$n$ patch $k$ is:
\begin{small}
\begin{align}
  \begin{pmatrix}
    x_i'\\
    y_i'\\
    z_i'\\
\end{pmatrix} =  \frac{\sum\limits_{m_i}^{M_d}\sum\limits_{n_i}^{N_d}(B_{M_d})_{m}^{m_i}(u)(B_{N_d})_{n}^{n_i}(v)\mathbf{c}_{{m_i}{n_i}}(c_{w})_{{m_i}{n_i}}w_{ik} }{\sum\limits_{m_i,n_i}(B_{M_d})_m^{m_i}(u)(B_{N_d})_n^{n_i}(v)(c_{w})_{{m_i}{n_i}}w_{ik}}, 
\end{align}
\end{small}
where $\mathbf{c}_{{m_i}{n_i}}$ denotes the control point coordinates and $(c_{w})_{{m_i}{n_i}}$ denotes its weight, and $w_{ik}$ is the element of $\mathbf{W}$. 

If we also input the normal $(n_{x_i}, n_{y_i}, n_{z_i})$ for point $p_i$, we can also reconstruct the normal $(n_{x_i}', n_{y_i}', n_{z_i}')$ by:
 \begin{align}
  \begin{pmatrix} n_{x_i}'\\ n_{y_i}'\\ n_{z_i}'\\ \end{pmatrix} =  
    \begin{pmatrix} \frac{\partial x_i'}{\partial u}\\\frac{\partial y_i'}{\partial u}\\ \frac{\partial z_i'}{\partial u}\\ \end{pmatrix} \times
    \begin{pmatrix} \frac{\partial x_i'}{\partial v}\\\frac{\partial y_i'}{\partial v}\\ \frac{\partial z_i'}{\partial v}\\ \end{pmatrix}, 
    \label{eq-recon-normal}
\end{align}
where $\times$ denotes the cross product.

${\mathbf{p}}_i$ denotes the input point coordinates.
${\mathbf{p}}_i^*$ denotes the reconstructed point coordinates. ${\mathbf{n}}_{p_i}$ denotes the input point normals.
${\mathbf{n}}_{p_i}^*$ denotes the reconstructed normals. The coordinate loss is defined as:
 \begin{align}
 L_{\text{coord}} = \frac{1}{N}\sum_{i=0}^{N-1}\|{\mathbf{p}}_i- {\mathbf{p}}_i^*\|_2^2.
\end{align}
If we also input the normals, the normal loss is defined as:
 \begin{align}
 L_{\text{norm}} = \frac{1}{N}\sum_{i=0}^{N-1}(1 - |{\mathbf{n}}_{p_i}^T{\mathbf{n}}_{p_i}^*|).
\end{align}

The loss for the reconstruction module is defined as: 
 \begin{align}
 L_{\text{recon}} = 
 \begin{cases}
 L_{\text{coord}}, & \mathrm{without \ normals},  \\
 L_{\text{coord}}+L_{\text{norm}}, & \mathrm{with \ normals}.
\end{cases} 
\end{align}

\subsubsection{Total Loss} 
The total loss is defined as the sum of decomposition, fitting, embedding, and reconstruction losses: $L = L_{\text{dec}} + L_{\text{fit}} + L_{\text{emb}} + L_{\text{recon}}$. We do not use different weights for each loss item because all point clouds are normalized into a unit sphere. Moreover, the $uv$ parameters are outputted directly from a $sigmoid$ layer, and the control points are outputted directly by a $tanh$ layer. Thus, each loss item is almost at the same scale, so we do not need different weights for each loss item. Furthermore, we use different learning rates for different modules to balance the training. Specific training details are listed in section \ref{training-details}.

\section{Experiments}

\subsection{Dataset Pre-Processing} 

We evaluate our approach on the ABC dataset \cite{Koch_2019_CVPR}. However, the ABC dataset does not have the annotations to learn B\'ezier decomposition on point clouds. Therefore, we do a pre-processing step. Specifically, we utilize the CGAL library \cite{cgal} and OpenCascade library \cite{opencascade} to perform B\'ezier decomposition on STEP files directly and perform random sampling on the surface to obtain the following labels: point coordinates, point normals, point $uv$ parameters, surface patch indices of the corresponding points, surface patch degrees, and surface patch control points. Finally, we use 5,200 CAD models for training and 1,300 CAD models for testing. Each CAD model contains randomly sampled $8,192$ points (non-uniform) with annotations. 

\subsection{Training Details}
\label{training-details}
We train a multi-task learning model. The learning rates differ depending on the MLP branch. The learning rate for the backbone, soft membership, and $uv$ parameters is set to $10^{-3}$, while the learning rate for the degree probabilities and control points is set to $10^{-4}$. As we have several learning tasks that are not independent, we set a lower learning rate for loss items, such as degree probabilities which converges faster. We set $\gamma$ as $3.0$ for the focal loss, and $\delta_{\text{pull}}$ as $0$ and $\delta_{\text{push}}$ as $2.0$ for the embedding losses. We employ ADAM to train our network. The model is then trained using $150$ epochs.

\subsection{Comparisons} 

We compare our algorithm with SPFN, ParSeNet, and HPNet \cite{li2019supervised,sharma2020parsenet,yan2021hpnet}.
We use both points and normals for training all the algorithms.
Since SPFN only supports four types of canonical primitives (plane, sphere, cone, and cylinder), we consider points belonging to other primitives falling out of the supported canonical primitive types as the ``unknown'' type. To make fair comparisons, we modify SPFN to let the network take point coordinates and normals as input for training. For ParSeNet, we only train the segmentation module on the ABC dataset. We use their pre-trained fitting model (SplineNet) directly. For HPNet, we also use the pre-trained fitting model directly, which is the same as ParSeNet. We observed that the output of HPNet is very sensitive to the number of points. In order to use HPNet at its best, we down-sample the point clouds to 7k points for training and testing.
We choose the following evaluation metrics:
\begin{enumerate}
\item \textbf{Primitive Type Accuracy} (``Acc''): \\ $\frac{1}{K}\sum_{k=0}^{K-1} \mathbb{I}(t_k==\hat{t}_k)$, where $t_k$ and $\hat{t}_k$ are predicted primitive type and ground truth type, respectively. This is used to measure the type accuracy. Note that our primitive types differ from other baselines.
\item \textbf{Rand Index} (``RI''): \\
$\frac{a+b}{c}$, where $c$ is $\binom{N}{2}$ denoting the total possible pairs for all points, and $a$ denotes the number of pairs of points that are both in the same primitive of prediction and ground truth, while $b$ denotes the number of pairs of points that are in a different primitive of prediction and ground truth. Rand index is a similarity measurement between two instances of data clustering, and a higher value means better performance \cite{chen2009benchmark,yi2018deep}.
\item \textbf{Normal Error} (``Err''): \\
$\frac{1}{N}\sum_{i=0}^{N-1} {\mathrm{arccos}}(
|{\mathbf{n}}_{p_i}^T{\mathbf{n}}_{p_i}^*|)$, where ${\mathbf{n}}_{p_i}$ and ${\mathbf{n}}_{p_i}^*$ are ground truth and predicted unit normal, respectively.
\item \textbf{Inference Time} (``Time''): \\
The inference time on the whole test dataset.
\item \textbf{Average Primitive Number} (``Num''): \\
The predicted average number of primitives on the whole test data set.
\end{enumerate}

\begin{figure*}
\centering
\subfigure[GT-Basic]{
    \begin{minipage}[b]{0.13\linewidth} 
    \includegraphics[width=1.8cm]{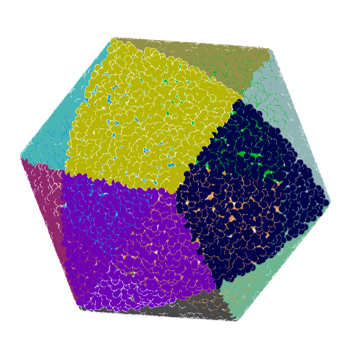}\vspace{0pt} 
    \includegraphics[width=1.8cm]{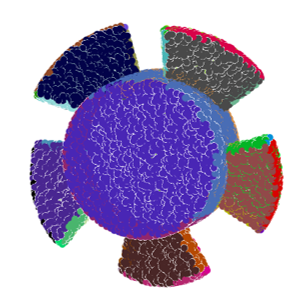}\vspace{0pt} 
    \includegraphics[width=1.8cm]{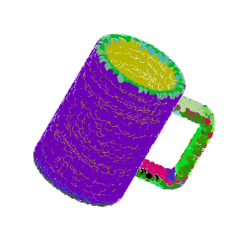}\vspace{0pt}
    \includegraphics[width=1.8cm]{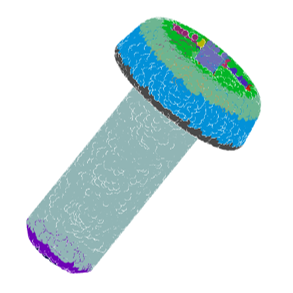}\vspace{0pt}
    \end{minipage}
}
\hspace{2mm}
\subfigure[HPNet]{
    \begin{minipage}[b]{0.13\linewidth} 
    \includegraphics[width=1.8cm]{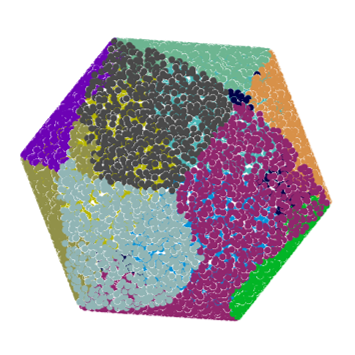}\vspace{0pt} 
    \includegraphics[width=1.8cm]{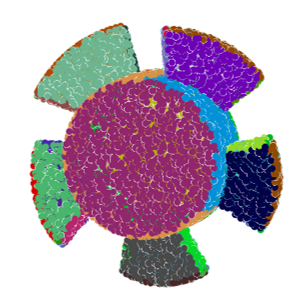}\vspace{0pt} 
    \includegraphics[width=1.8cm]{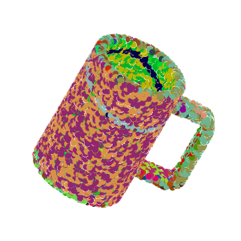}\vspace{0pt}
    \includegraphics[width=1.8cm]{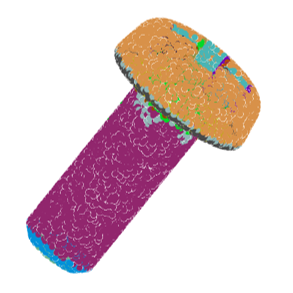}\vspace{0pt}
    \end{minipage}
}
\hspace{2mm}
\subfigure[ParSeNet]{
    \begin{minipage}[b]{0.13\linewidth} 
    \includegraphics[width=1.8cm]{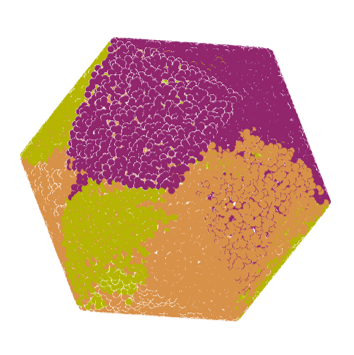}\vspace{0pt} 
    \includegraphics[width=1.8cm]{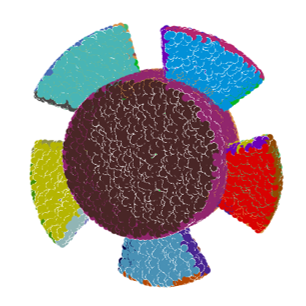}\vspace{0pt} 
    \includegraphics[width=1.8cm]{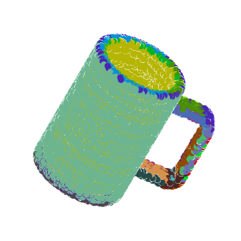}\vspace{0pt}
    \includegraphics[width=1.8cm]{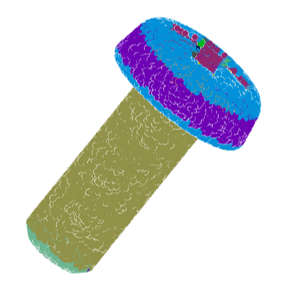}\vspace{0pt}
    \end{minipage}
}
\hspace{2mm}
\subfigure[SPFN]{
    \begin{minipage}[b]{0.13\linewidth} 
    \includegraphics[width=1.8cm]{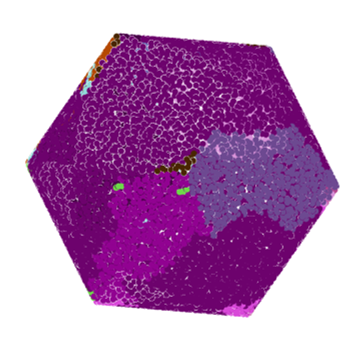}\vspace{0pt} 
    \includegraphics[width=1.8cm]{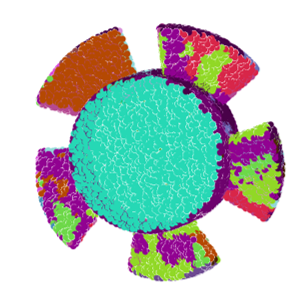}\vspace{0pt} 
    \includegraphics[width=1.8cm]{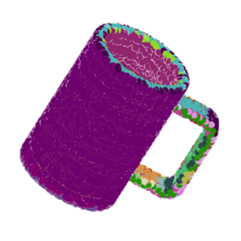}\vspace{0pt}
    \includegraphics[width=1.8cm]{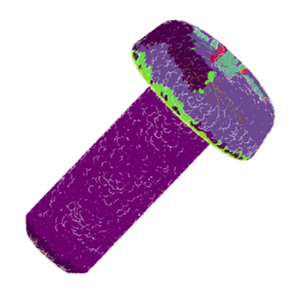}\vspace{0pt}
    \end{minipage}
}
\hspace{2mm}
 \subfigure[GT-Ours]{
    \begin{minipage}[b]{0.13\linewidth} 
    \includegraphics[width=1.8cm]{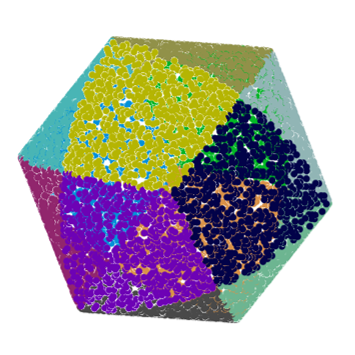}\vspace{0pt} 
    \includegraphics[width=1.8cm]{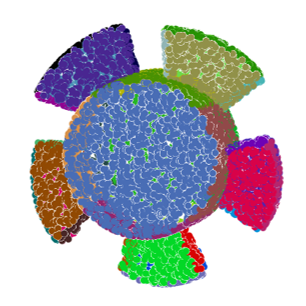}\vspace{0pt} 
    \includegraphics[width=1.8cm]{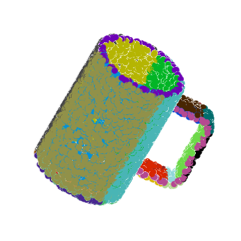}\vspace{0pt}
    \includegraphics[width=1.8cm]{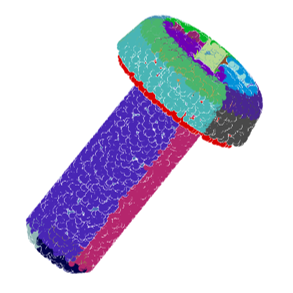}\vspace{0pt}
    \end{minipage}
}
\hspace{2mm}
\subfigure[Ours]{
    \begin{minipage}[b]{0.13\linewidth} 
    \includegraphics[width=1.8cm]{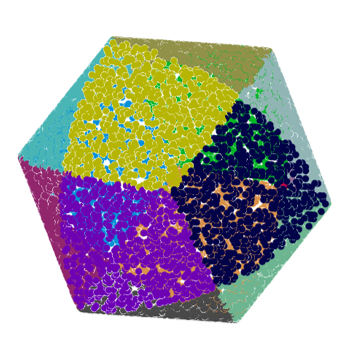}\vspace{0pt} 
    \includegraphics[width=1.8cm]{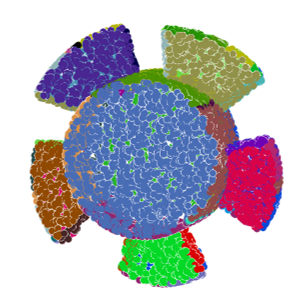}\vspace{0pt} 
    \includegraphics[width=1.8cm]{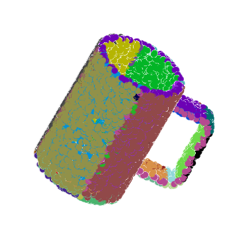}\vspace{0pt}
    %1.0\columnwidth
    \includegraphics[width=1.8cm]{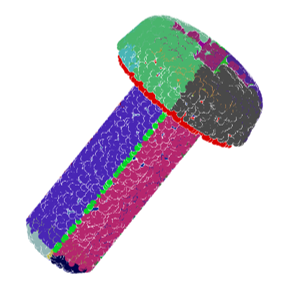}\vspace{0pt}
    \end{minipage}
}
\caption{Comparisons on the ABC dataset, where GT-Ours denotes the ground truth for B\'ezier decomposition, and GT-Basic denotes the ground truth for HPNet, ParSeNet, and SPFN.}
\label{comparisons}
\end{figure*}

We record these evaluation metrics in table \ref{table-comparisons} and \ref{table-comparisons-without-inputting-normals}. Figure \ref{comparisons} shows visual depictions of the results. Our results show the best performance regarding primitive type accuracy, normal fitting error, and inference time. Our method is much faster for inference because it uses a general formula for different primitive types, and the embedding module is free of iterations. Other methods treat primitives with different equations, and ParSeNet and HPNet need a mean-shift step. Even though our approach may lead to more segmented primitives by the nature of B\'ezier decomposition, the evaluation metrics of primitive type accuracy and normal fitting error are computed in a point-wise manner. Thus, over-segmentation and under-segmentation will not lead to smaller or bigger errors due to fewer or more segmented primitives. 

\begin{table}
\begin{center}
\resizebox{1.0\columnwidth}{!}{
\begin{tabular}{l| ccccc}
\toprule
\multirow{1}{*}{Method} &\multirow{1}{*}{Acc(\%)($\uparrow$)} &\multirow{1}{*}{RI(\%)($\uparrow$)} &\multirow{1}{*}{Err(rad)($\downarrow$)} &\multirow{1}{*}{Time(min)($\downarrow$)} &\multirow{1}{*}{Num}\\
\midrule
\multirow{1}*{HPNet}
& 94.09 & 97.76 & 0.1429 & 1120.31 & 13.75 \\
\multirow{1}*{ParSeNet}
& 94.98 & 97.18 & -  & \enspace252.01  & 14.06 \\
\multirow{1}*{SPFN}
& 83.20 & 93.03 & 0.1452 & \quad11.52   & 21.02\\
\multirow{1}*{Ours}
& \textbf{96.83} & 95.68 & \textbf{0.0522} & \enspace\quad\textbf{4.25} & 19.17\\
\bottomrule
\end{tabular}}
\end{center}
\caption{Evaluation for primitive instance segmentation on the ABC-decomposition dataset. ``rad'' denotes the radian. Here we input both point coordinates and normals.}
\label{table-comparisons}
\end{table}

We also show the performance of all the methods without normals as input. For our method and SPFN, we only input point coordinates into the neural networks but use normals as supervision. Since ParSeNet does not regress normals, we cannot use normals as supervision. We train ParSeNet without normals as input to test its performance. HPNet uses the network to regress the normals from the input and also utilizes the ground truth normals to construct an affinity matrix as a post-processing step for clustering. We modify HPNet to let the affinity matrix be constructed from the regressed normals instead of the ground-truth normals. Table \ref{table-comparisons-without-inputting-normals} records the evaluation metrics of each method. From the experiments, we deduce that normals are important for the task of parametric primitive segmentation.

\begin{table}
\begin{center}
\resizebox{1.0\columnwidth}{!}{
\begin{tabular}{l| ccccc}
\toprule
\multirow{1}{*}{Method}  &\multirow{1}{*}{Acc(\%)($\uparrow$)} &\multirow{1}{*}{RI(\%)($\uparrow$)} &\multirow{1}{*}{Err(rad)($\downarrow$)} &\multirow{1}{*}{Time(min)($\downarrow$)} &\multirow{1}{*}{Num}\\
\midrule
\multirow{1}*{HPNet}
& 90.45 & 96.04 & 0.2256 & 1165.95 & 22.08 \\
\multirow{1}*{ParSeNet}
& 90.31 & 94.55 & -   & \enspace206.74  & 15.39 \\
\multirow{1}*{SPFN}
& 72.90 & 80.76 & 0.3309 & \quad10.46 & 27.61\\
\multirow{1}*{Ours}
& \textbf{92.34} & 90.54 & \textbf{0.2003} & \enspace\quad\textbf{4.05} & 33.67\\
\bottomrule
\end{tabular}}
\end{center}
\caption{Here we only input point coordinates into the neural network.}
\label{table-comparisons-without-inputting-normals}
\end{table}

\subsection{Ablation Studies}

\begin{figure}[ht]
\centering
\includegraphics[height=5.5cm]{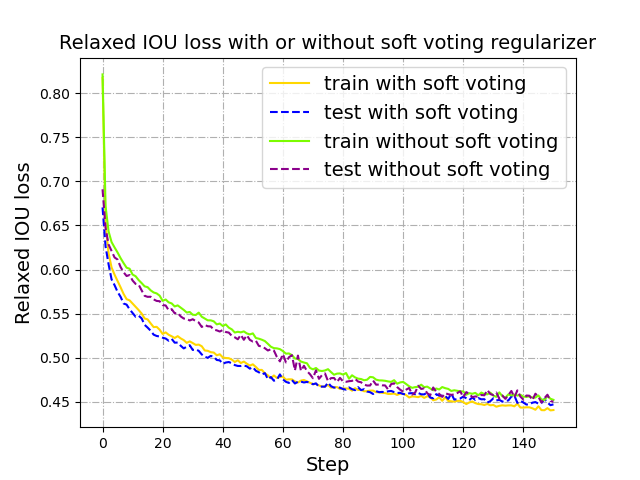}
\caption{Relaxed IOU loss with or without soft voting loss.}
\label{loss-riou}
\end{figure}

We first conduct experiments to verify the usefulness of the soft voting regularizer. The soft voting regularizer favors point primitive type consistency for each primitive instance, i.e., points assigned to the same primitive instance should have the same primitive type. From our experiment, we find that the soft voting regularizer not only improves the primitive type accuracy but also accelerates training relaxed IOU. Please refer to figure \ref{loss-riou} and the last two rows of table \ref{table-ablation}.

\begin{table}
\begin{center}
\resizebox{0.85\columnwidth}{!}{
\begin{tabular}{l|cccc}
\toprule
\multirow{1}{*}{Module}  &\multirow{1}{*}{Acc(\%)($\uparrow$)} &\multirow{1}{*}{RI(\%)($\uparrow$)} &\multirow{1}{*}{Err(rad)($\downarrow$)}  &\multirow{1}{*}{Num} \\
\midrule
\multirow{1}*{D} 
 & 97$+$0.10 & 96$+$0.83 & 1.0982 & 24.46 \\
\multirow{1}*{D+E}
& 97$+$0.26 & 96$-$0.36 & 0.9834 & 22.54 \\ %24.54 
\multirow{1}*{D+F}
& 97$+$0.19 & 96$+$0.52 & 0.5424 & 23.45 \\
\multirow{1}*{D+E+F} 
 & 97$-$0.02 & 96$-$0.39 & 0.4884 & 20.85 \\
\multirow{1}*{D+F+R}
 & 97$+$0.19 & 96$+$0.48 & 0.0819 & 23.08 \\
 \multirow{1}*{No-voting}
 & 97$-$1.32 & 96$-$0.44 & 0.0547 & 19.45 \\
\multirow{1}*{Full-module}
 & 97$-$0.17 & 96$-$0.32 & 0.0522 & 19.17 \\
\bottomrule
\end{tabular}}
\end{center}
\caption{Ablation study. D denotes the decomposition module. E denotes the embedding module. F denotes the fitting module. R denotes the reconstruction module. No-voting denotes using all modules without the soft voting regularizer. Full-module denotes using all the modules plus the soft voting regularizer.}
\label{table-ablation}
\end{table}

We also verify the functionalities of each module. If we only use the decomposition module, the result is not good even though the ``Acc'' and ``RI'' are slightly higher because the decomposition module ignores the fitting, limiting the segmentation applicable to specific datasets. The reconstruction module reduces the ``Err'' significantly compared to the fitting module because the reconstruction module controls how ``well-fitted'' a predicted B\'ezier primitive is to the input point clouds. In contrast, the fitting module only regresses the control points and $uv$ parameters. The embedding module is designed to eliminate small patches that contain few points, seeing the ``Num'' column. Therefore, experimenting with the embedding module results in fewer patch numbers than its counterpart. To conclude, training with all the modules yields the best results.

\subsection{Stress Tests}

\begin{figure}
\centering
\subfigure[HPNet]{
    \begin{minipage}[b]{0.18\linewidth} 
    \includegraphics[width=1.0\columnwidth]{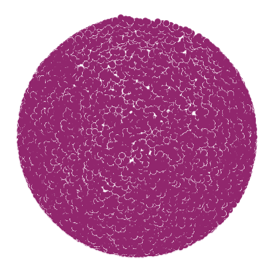}
    \includegraphics[width=1.0\columnwidth]{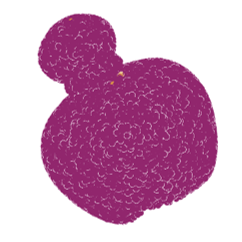}
    \includegraphics[width=1.0\columnwidth]{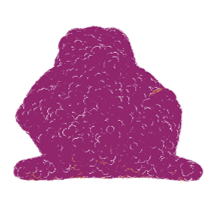}
    \includegraphics[width=1.0\columnwidth]{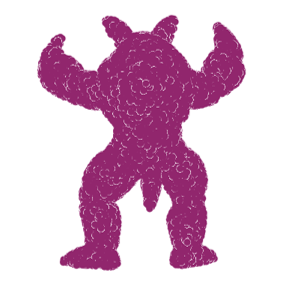}
    \end{minipage}
}
\hspace{2mm}
\subfigure[ParSeNet]{
    \begin{minipage}[b]{0.18\linewidth} 
    \includegraphics[width=1.0\columnwidth]{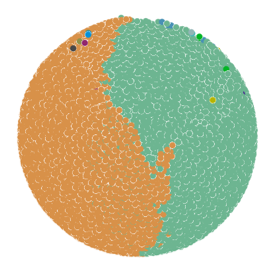}
    \includegraphics[width=1.0\columnwidth]{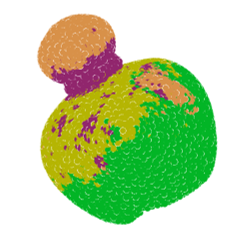}
    \includegraphics[width=1.0\columnwidth]{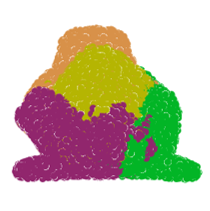}
    \includegraphics[width=1.0\columnwidth]{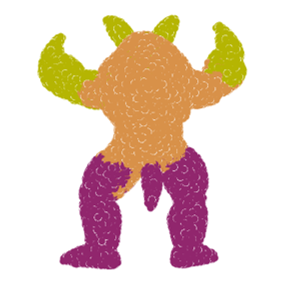}\vspace{0pt}
    \end{minipage}
}
\hspace{2mm}
\subfigure[SPFN]{
    \begin{minipage}[b]{0.18\linewidth} 
    \includegraphics[width=1.0\columnwidth]{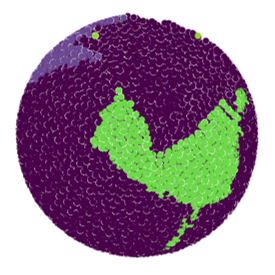}
    \includegraphics[width=1.0\columnwidth]{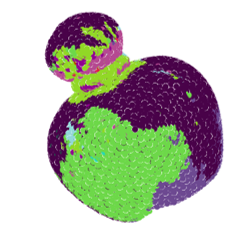}
    \includegraphics[width=1.0\columnwidth]{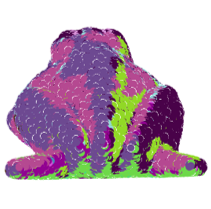} 
    \includegraphics[width=1.0\columnwidth]{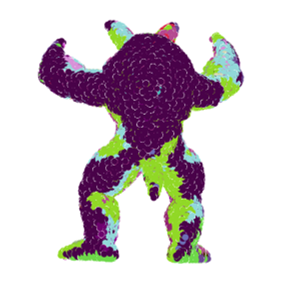}\vspace{0pt}
    \end{minipage}
}
\hspace{2mm}
\subfigure[Ours]{
    \begin{minipage}[b]{0.18\linewidth} 
    \includegraphics[width=1.0\columnwidth]{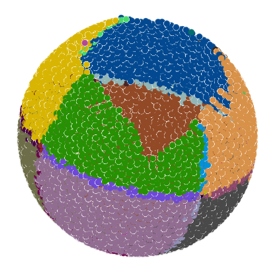}
    \includegraphics[width=1.0\columnwidth]{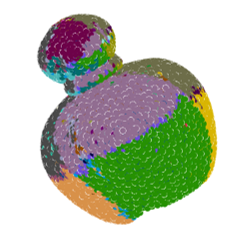}
    \includegraphics[width=1.0\columnwidth]{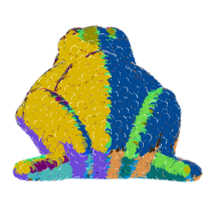}
    \includegraphics[width=1.0\columnwidth]{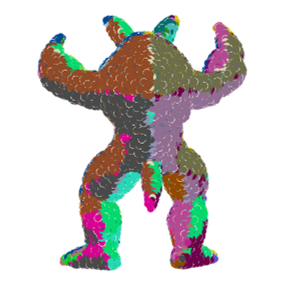}
    \end{minipage}
}
\caption{Stress tests on real-scan data. The above two rows are CAD point clouds, and the last two are free-form point clouds.}
\label{stress-test}
\end{figure}

To test whether our algorithm can work in real-world scenarios, we show more results from the real-scan data from the Aim@Shape dataset \cite{falcidieno2004aim}. The sampling is non-uniform, with missing data and measurement noise compared to the ABC dataset. Besides, We cannot train the network on those data directly because they lack ground-truth labels. Instead, we use the models trained on the ABC dataset and test the performance on real-scan data. Our algorithm still works, while other methods are sensitive. Another positive aspect is that our algorithm could decompose the axis-symmetric free-form point clouds with much smoother boundaries of different patches. Please refer to figure \ref{stress-test}. 

We also test the performance of our network by adding Gaussian white noise. Specifically, we apply different scales of Gaussian white noise to the point coordinates after normalizing them into a unit sphere. The noise scale denotes the standard deviation of the Gaussian white noise. It ranges from $0.01$ to $0.05$. We train our network on noise-free data but test the network with Gaussian white noise. Please refer to table \ref{table-noise}.

\begin{table}
\begin{center}
\resizebox{0.85\columnwidth}{!}{
\begin{tabular}{l|cccc}
\toprule
\multirow{1}{*}{Noise scale}  &\multirow{1}{*}{Acc(\%)($\uparrow$)} &\multirow{1}{*}{RI(\%)($\uparrow$)} &\multirow{1}{*}{Err(rad)($\downarrow$)}  &\multirow{1}{*}{Num} \\
\midrule
\multirow{1}*{No-noise} 
& \textbf{96.83} & \textbf{95.68} & \textbf{0.0522} & 19.17 \\
\multirow{1}*{0.01}
& 96.75 & 94.27 & 0.0525 & 20.38 \\
\multirow{1}*{0.02}
& 96.63 & 93.48  & 0.0529 & 21.68 \\
\multirow{1}*{0.03}
& 96.34 & 92.73 & 0.0538 & 22.76 \\
\multirow{1}*{0.04}
& 96.15 & 92.04 & 0.0552 & 23.68 \\
\multirow{1}*{0.05}
& 96.07 & 91.34 & 0.0559 & 24.38 \\
\bottomrule
\end{tabular}}
\end{center}
\caption{Evaluation of our algorithm at different noise scales.}
\label{table-noise}
\end{table}

\subsection{Applications}

\begin{figure}
\centering
\subfigure{
\begin{minipage}[b]{0.05\linewidth}
\centering
\rotatebox{90}{ \footnotesize{\quad \quad ParSeNet}}
\rotatebox{90}{ \footnotesize{\quad \quad   Ours}}
\end{minipage}
}
\subfigure{
    \begin{minipage}[b]{0.2\linewidth}
    \centering
    \includegraphics[width=1.8cm]{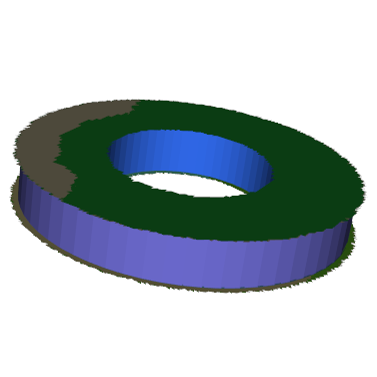}\vspace{0pt} 
    \includegraphics[width=1.8cm]{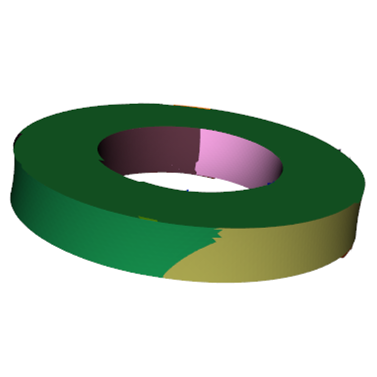}\vspace{0pt} 
    \end{minipage}
}
\subfigure{
    \begin{minipage}[b]{0.2\linewidth}
    \centering
    \includegraphics[width=1.8cm]{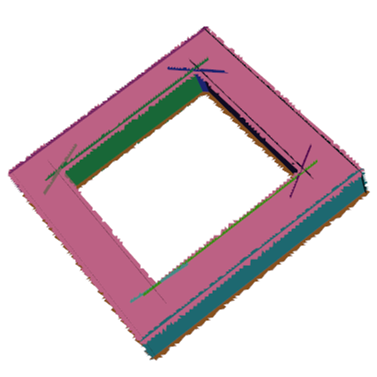}\vspace{0pt} 
    \includegraphics[width=1.8cm]{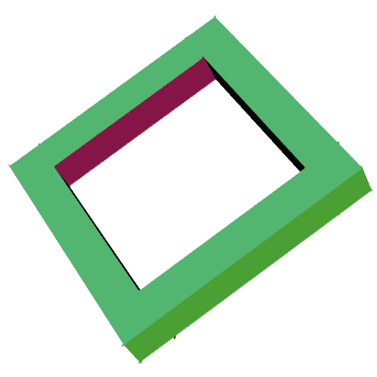}\vspace{0pt} 
    \end{minipage}
}
\subfigure{
    \begin{minipage}[b]{0.2\linewidth} 
    \centering
    \includegraphics[width=1.8cm]{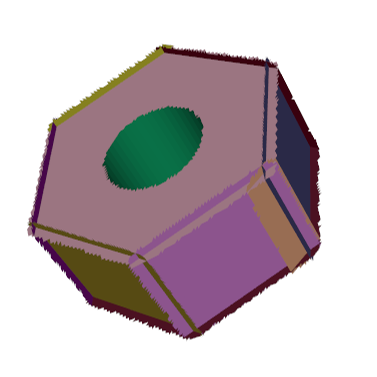}\vspace{0pt} 
    \includegraphics[width=1.8cm]{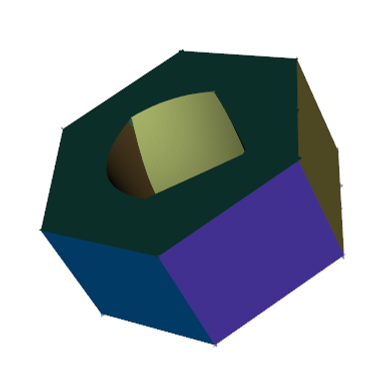}\vspace{0pt} 
    \end{minipage}
}
\subfigure{
    \begin{minipage}[b]{0.2\linewidth} 
    \centering
    \includegraphics[width=1.8cm]{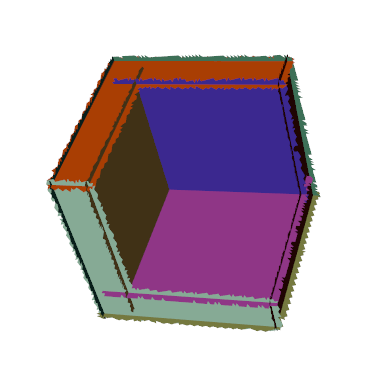}\vspace{0pt} 
    \includegraphics[width=1.8cm]{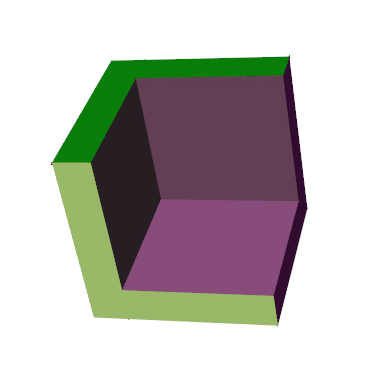}\vspace{0pt} 
    \end{minipage}
}
\caption{Reconstruction of the full B\'ezier model and visual comparison with ParSeNet.}
\label{refit-bezier}
\end{figure}

We can reconstruct the full B\'ezier model from the B\'ezier primitive segmentation. We do not follow ParSeNet to pre-train a model that outputs a fixed control point size. Instead, we reuse the rational B\'ezier patch to refit the canonical B\'ezier patch. We treat the degrees of the canonical B\'ezier patch the same as the rational B\'ezier patch. As a result, we fetch the segmentation and degrees of each patch predicted from the network. Then, we use the parameterization \cite{levy2002least} to recompute $uv$ parameters and least squares to refit control points for each patch. Each patch is expanded by enlarging the $uv$ domain to guarantee intersections with its adjacent patches. After that, we use the  CGAL co-refinement package \cite{cgal} to detect intersecting polylines for adjacent tessellated patches and trim the tessellated patch with the intersected polylines. Our reconstructed full B\'ezier model can preserve the sharp features, while the boundaries of ParSeNet for different primitives are jaggy and thus fail to preserve the sharp features. Please refer to figure \ref{refit-bezier}.

\section{Conclusion}

This paper presents an end-to-end method to group points by learning B\'ezier decomposition. In contrast to approaches treating different geometric primitives separately, our method uses a general formulation for different primitive types. Regarding limitations, B\'ezier decomposition may naturally generate overly complex segmentations. In addition, we choose the rational B\'ezier patch as the primitive type. As the formulation is not linear, fitting the parametric patch is not direct. In future work, we wish to use the neural network to directly regress the canonical B\'ezier patch.

\section*{Acknowledgements}
This research is part of a project that has received funding from the European Union’s Horizon 2020 research and innovation program under the Marie Skłodowska-Curie grant agreement No. 860843. The work of Pierre Alliez is also supported by the French government, through the 3IA Côte d'Azur Investments in the Future project managed by the National Research Agency (ANR) with the reference number ANR-19-P3IA-0002. 

\bibliographystyle{named}
\bibliography{ijcai23}

\end{document}